\documentclass[conference]{IEEEtran}
\usepackage{cite}
\usepackage{amsmath,amssymb,amsfonts}
\usepackage{algorithmic}
\usepackage{graphicx}
\usepackage{textcomp}
\usepackage{xcolor}
\def\BibTeX{{\rm B\kern-.05em{\sc i\kern-.025em b}\kern-.08em
    T\kern-.1667em\lower.7ex\hbox{E}\kern-.125emX}}

\usepackage{bm}
\usepackage{acro}

\acsetup{}
\DeclareAcronym{rl}{
short = RL,
long = Reinforcement Learning 
}
\DeclareAcronym{tm}{
short = transition model,
long = transition model,
plural = s
}

\begin{document}

\title{Relate to Predict: Towards Task-Independent Knowledge Representations for Reinforcement Learning
}

\author{

\IEEEauthorblockN{1\textsuperscript{st} Thomas Schnürer}
\IEEEauthorblockA{
\textit{Ilmenau University of Technology}\\
Ilmenau, Germany \\
thomas.schnuerer@tu-ilmenau.de}
\and

\IEEEauthorblockN{2\textsuperscript{nd} Malte Probst}
\IEEEauthorblockA{
\textit{Honda Research Institute EU}\\
Offenbach, Germany \\
malte.probst@honda-ri.de} 
\and

\IEEEauthorblockN{3\textsuperscript{rd} Horst-Michael Gross}
\textit{Ilmenau University of Technology}\\
Ilmenau, Germany \\
horst-michael.gross@tu-ilmenau.de
}

\maketitle

\begin{abstract}
Reinforcement Learning (RL) can enable agents to learn complex tasks. However, it is difficult to interpret the knowledge and reuse it across tasks. Inductive biases can address such issues by explicitly providing generic yet useful decomposition that is otherwise difficult or expensive to learn implicitly. For example, object-centered approaches decompose a high dimensional observation into individual objects. Expanding on this, we utilize an inductive bias for explicit object-centered knowledge separation that provides further decomposition into semantic representations and dynamics knowledge. For this, we introduce a semantic module that predicts an objects' semantic state based on its context. The resulting affordance-like object state can then be used to enrich perceptual object representations. With a minimal setup and an environment that enables puzzle-like tasks, we demonstrate the feasibility and benefits of this approach. Specifically, we compare three different methods of integrating semantic representations into a model-based RL architecture. Our experiments show that the degree of explicitness in knowledge separation correlates with faster learning, better accuracy, better generalization, and better interpretability.
\end{abstract}


\section{Introduction}

\acf{rl} is a very powerful approach that enables agents to successfully learn well-specified tasks. 
However, it usually leads to trained models that are also bound to a single task, making it difficult to reuse the expensively learned knowledge. 
One approach to overcome this limitation is the introduction of inductive biases that encourage a purposeful decomposition. For example, separating knowledge about tasks from generic knowledge that can be reused across tasks \cite{Vehicle2018}.
Additionally, recent approaches have highlighted many advantages of object-centered decomposition \cite{Garnelo2019,Anon2019,Locatello2020}.
Following this trend, we understand knowledge about objects and their relations as one form of generic knowledge. For example, interactions with objects (pick up), semantic relations (key and lock) and affordances (can be pushed) can be reused across many tasks, whereas only their role in regards to a specific goal might differ. Such an object-centered separation might also be one of the reasons why humans are able to acquire new tasks quickly \cite{Dubey2018}.

In a similar fashion, we further divide object-centered knowledge into semantic and dynamics knowledge. While dynamics involves fine grained step-wise interactions (e.g. physical manipulation), semantics concerns discrete properties that are consistent over a long period of time steps and can influence dynamics.
For example, for a robot learning to sort objects into containers, dynamics might include the physical object manipulations while semantics involves whether certain objects fit into specific containers. 

In this paper, we propose an object-centered inductive bias for model-based \ac{rl} that explicitly separates semantic and dynamics knowledge. We deploy this inductive bias with an architecture that enriches perceptual object representations with semantic information that is derived from context.
Building upon recent advances with task-free object-based models \cite{Watters2019}, we demonstrate three possible implementations of such a separation with varying degrees of explicitness and analyze their performance on an environment in which complex puzzle like tasks can be created. Our experiments show that the degree of explicitness correlates with several desirable qualities.

\section{State of the Art}
In model-free \ac{rl}, the state space is always task-specific by design, therefore limiting the reusability of results. 
Transfer learning comprises a multitude of domains that try to overcome such limitations \cite{Taylor2009}.
Aiming for a transfer to arbitrary tasks that are not known during training, model-based \ac{rl} can be a means to learn task-independent knowledge. There, a \acs{tm} is learned as an approximation of the real environment, which, in turn, can be used as a surrogate to train or plan for novel tasks later on \cite{Ha2018,Moerland}. 
However, in complex environments where an exhaustive exploration of all possible world states is not feasible, learning a sufficiently accurate TM is very challenging. Addressing such challenges, recent approaches have demonstrated various benefits of a state decomposition, e.g. into locally independent regions \cite{Pitis} or into individual entities \cite{Anon2019,Veerapaneni,Keramati}. Intuitively, such an inductive bias helps to find structure in the data (e.g.\ the existence of individual objects) that is otherwise expensive or difficult to learn. This can be a crucial advantage in complex environments because it allows learning about entities independent of the situation and apply this knowledge to new situations \cite{Zambaldi2019}.

These advances also seem to be in line with the inductive biases that humans leverage in order to learn new tasks quickly. According to \cite{Dubey2018}, human biases can be separated into three stages: first, the concepts of objects and visual similarity are learned. Second, semantics and affordances of objects can be aquired. Finally, object interactions (dynamics) are attained.

In this regard, state decomposition into objects can be seen as analog to the first stage and is already addressed with a multitude of approaches, including simple CNNs \cite{Zambaldi2019} and VAEs \cite{Burgess}. However, the other stages are usually not considered separately. In an effort to address this separation, we introduce an inductive bias that can enrich the perceptual representation of objects with semantic information. In this paper, we specifically focus on relational semantic knowledge, arising from the relations of one object to other objects.

\section{Model}
\label{sec:model}

We assume the existence of an object extractor $V_{enc}$ that decomposes an observation $\bm{x} \in\mathbb{R}^n$ (i.e.\ an image) into a scene 

\begin{equation}
\bm{z} \in \mathbb{R}^{K\times M} = V_{enc}( \bm{x} )
\end{equation}

 so that a scene $\bm{z} = \{z^1, z^2, ..., z^K\}$ contains $K$ objects of dimension $M$ (e.g.\ \cite{Burgess}). Consequently, the state of an object $z^k$ in a scene is purely defined by perceptual features, i.e\ its directly observable properties (e.g.\ position, shape and color).
For simplification, we assume both $K$ and $M$ to be constant, but allow for slots to be empty with $z^k = [0]^M$.
In model-based RL, a transition model $TM$ maps the current state $\bm{z_{t}}$ and an action $a_t$
to a predicted next-step state 

\begin{equation}
\bm{\widetilde{z}_{t+1}} = TM(\bm{z_{t}}, a_t), 
\end{equation}

approximating the dynamics of the environment. With an object-centered state $\bm{z} = \{z^1, z^2, ..., z^K\}$, this usually concerns the behavior of objects for possible actions.
In the most simple case where objects do not interact with each other, a shared object-wise transition model $T_{Obj}$ is sufficient. Such a model predicts the future state $\bm{z}$ for all $K$ objects independently with 

\begin{equation}
\bm{\widetilde{z}^k_{t+1}} = T_{Obj}(\bm{z^k_t}, a_t).
\end{equation}

To also account for object interactions (such as physics), a number of dynamic transition model architectures \cite{Anon2019,Veerapaneni,Keramati} consider all objects simultaneously instead with  

\begin{equation}
\bm{\widetilde{z}_{t+1}} = T_{Dyn}(\bm{z_t}, a_t).
\end{equation}

\begin{figure}
\begin{center}
      \includegraphics[width=0.9\linewidth]{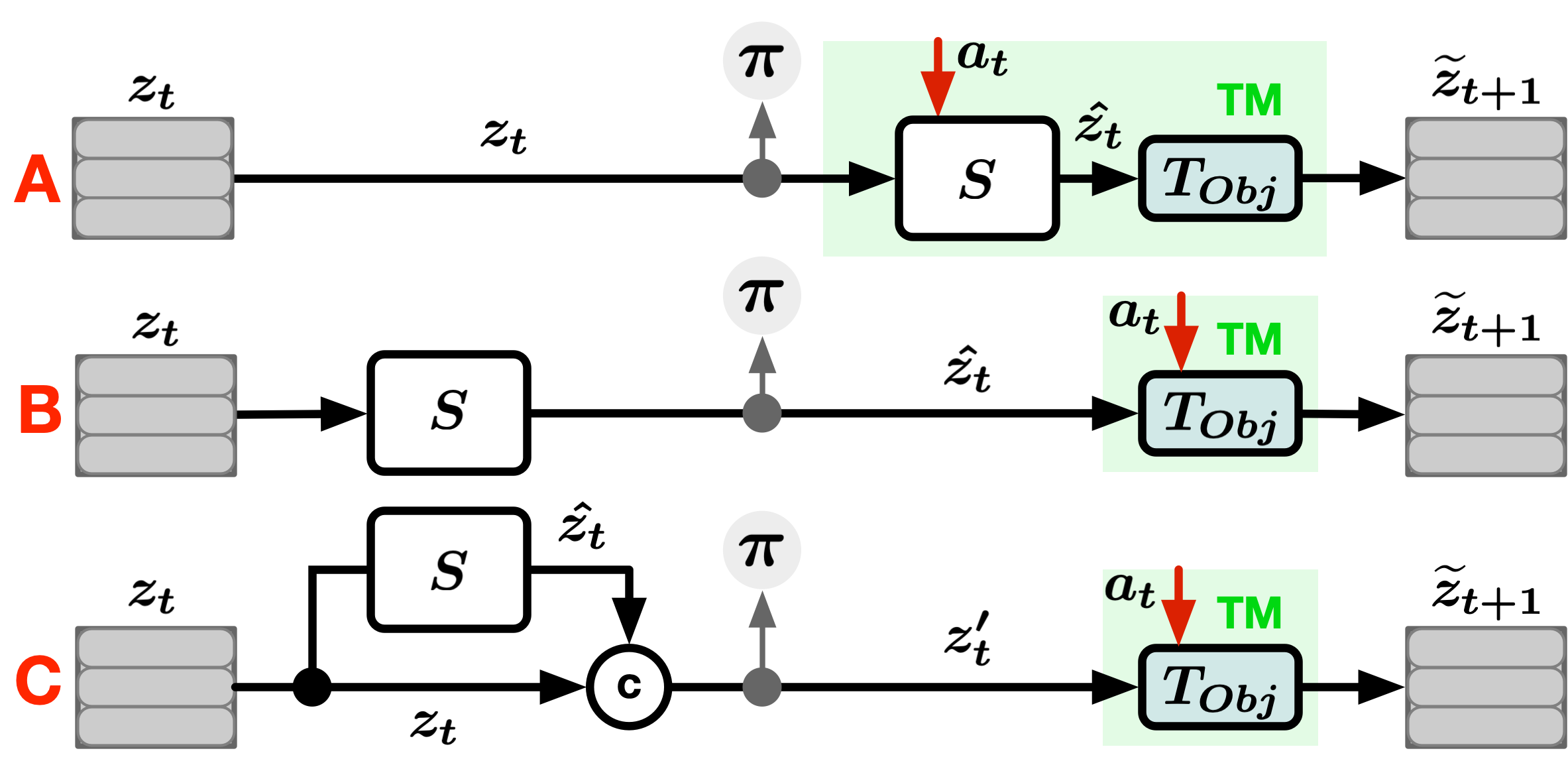}
\end{center}
   \caption{Implementation of the proposed inductive bias: three ways of integrating a semantic context module $S$ with an object-wise transition model $T_{Obj}$. The arrow towards $\pi$ indicates the representation that can be used to train a task policy on the transition model later on.
   \textbf{A (Internal):} The module for semantic context is part of transition model (similar to SotA).
   \textbf{B (Sequential):} Preceding the transition model, initial object features are transformed to include semantic information. 
   \textbf{C (Parallel):} Preceding the transition model, semantic information is concatenated to the initial object features.
   B and C allows the transition model (TM) to stay purely object-wise.}
\label{fig:model}
\end{figure}

These models usually consider both independent object dynamics and inter-object dynamics internally. A simple version of such a model can be seen in Fig. \ref{fig:model} (A).

Instead of encapsulating all relations between objects within the \ac{tm}, we propose an inductive bias that further decomposes the knowledge of \acp{tm}. Specifically, we introduce a semantic context module that can be independent of a \ac{tm}, explicitly separating semantic knowledge (such as semantic object states) from dynamics knowledge (such as object specific and inter-object dynamics). 

We expect the following benefits for such an approach:
\begin{itemize}
\item \textbf{Model learning}: Purposeful decomposition explicitly imposes structure into the data that otherwise needs to be learned implicitly. While this is already a benefit in itself, it additionally might allow a reduction in complexity of the individual modules. Both aspects can lead to higher data efficiency and shorter training times

\item \textbf{Generalization:} Similar to how an explicit decomposition into objects can greatly improve generalization, we expect an explicit knowledge decomposition to be beneficial when generalizing to new tasks or environments.

\item \textbf{Transparency and interpretability:} If the semantic information is explicitly represented, it is easier to understand the model's knowledge and behavior.

\item \textbf{Task learning}: Concepts learned by the \ac{tm} (e.g.\ spatial closeness, similarity of shapes and colors) can be directly used by a downstream task policy later on if they are explicitly exposed and disentangled from dynamics. This could simplify a policies state space substantially, allowing faster training on new tasks or even unlock tasks that might have been too complex otherwise.

\end{itemize}

In Section \ref{sec:experiments}, we examine the first three hypotheses empirically to substantiate the feasibility of such an approach and leave the last hypothesis for future work.

\subsection*{Semantic Module}

We define the semantic context module $S$  as a function 

\begin{equation}
f_{S} : \mathbb{R}^{K\times M} \to \mathbb{R}^{K\times N}
\end{equation}

that maps a list of objects $\bm{z} = \{z^1, z^2, ..., z^K\}$ to a list of semantic descriptions $\bm{\hat{z}} = \{\hat{z}^1, \hat{z}^2, ..., \hat{z}^K\}$ where each description $\hat{z}^k$ is a vector of length $N$, comprising semantic properties of object $z^k$ based on its context $C_{z^k}$.

\begin{equation}
\hat{z}^k = f(z^k, C_{z^k})
\end{equation}

For simplicity, we define the entire observation as context for all objects with $C_{z^k} = \bm{z},  \forall k$. However, other possibilities might include spatial constraints (e.g.\ proximity to the object or agent), temporal constraints (e.g.\ previous time steps or predictions), and other modalities (e.g. sound, environment properties, the agents' intrinsic state values or even control inputs).

We compare three different methods (internal, sequential and parallel, see Fig. \ref{fig:model}) of integrating $S$ into an architecture with an object-wise \ac{tm} $T_{Obj}$.

\begin{equation}
\bm{\widetilde{z}_{t+1}} = f(S, T_{Obj}, \bm{z}_t, a_t).
\end{equation}

The variants differ in their degree of explicitness regarding the separation between perceptual, semantic and dynamics knowledge. We hypothesize that the degree of explicitness correlates with the previously described benefits, especially for training a task policy $\pi$ on the \ac{tm} later on:

The \textbf{internal model (A)} combines $S$ and $T_{Obj}$ to form a simple version of the preciously discussed dynamic \acp{tm}:

\begin{equation}
\bm{\hat{z}_t} = S( \bm{z}_t, a_t), \qquad \bm{\widetilde{z}^k_{t+1}} = T_{Obj}(\bm{\hat{z}}^k_t)
\end{equation}

By design, $S$ has to consider interactions regarding action $a_t$. Consequently, the semantic information in $\bm{\hat{z}_t}$ is entangled with both the dynamics model and perceptual features. Therefore, it cannot be used for downstream modules (e.g. policy $\pi$). Also, the learned semantic information might be action specific.

In the \textbf{sequential model (B)}, perceptual features are transformed to also include semantic information:

\begin{equation}
\bm{\hat{z}}_t = S( \bm{z}_t), \qquad \bm{\widetilde{z}}^k_{t+1} = T_{Obj}(\bm{\hat{z}}^k_t, a_t)
\end{equation}

While semantic information within $\hat{\bm{z}}_t$ can now be used by downstream modules, it is still entangled with perceptual object representations $\bm{z}_t$. Given that the object extractor $V_{enc}$ produces disentangled, interpretable embeddings, these will be projected into a different embedding space and therefore might lose disentanglement and easy interpretability. Additionally, some features of the perceptual embedding $\bm{z}_t$ could get lost if they are not essential for the transition model $T_{Obj}$ to predict the next state.

Finally, the \textbf{parallel model (C)} preserves all features of the perceptual embedding $\bm{z}_t$ by concatenating them to the explicitly embedded semantic information $\hat{\bm{z}}_t$:

\begin{equation}
\bm{\hat{z}}_t = S( \bm{z}_t), \quad \bm{z}'_t = [\bm{z}_t, \bm{\hat{z}}_t],  \quad \bm{\widetilde{z}}^k_{t+1} = T_{Obj}(\bm{z}'^{k}_t, a_t)
\end{equation}
While this makes both of them easier to interpret, downstream modules might also benefit from this disentanglement during training. Importantly, $S$ can also be applied to the predicted next state $z_{t+1}$ for models B and C, allowing a task policy $\pi$ to learn in a semantically enriched state space.

\subsection*{Implementation}
The purpose of the semantic module is to describe an objects' semantic state based on its context. As a proof of concept, we define the context as the relation of an object to the environment. In an object-centric environment, this can be described by all dyadic object-wise relations of an object (including itself). Therefore, we implemented the semantics module $S$  as a three layer relation network \cite{Santoro2017} (64 features per layer) and used a three layer MLP (512 features per layer) for the object-wise transition model $T_{Obj}$. Both modules are followed by a linear layer each, transforming their output to the target dimension.

\section{Experiments}
\label{sec:experiments}

We now examine the feasibility and benefits of an explicit decomposition into semantic and dynamic knowledge. Specifically, we compare all three models (see Fig. \ref{fig:model}) regarding their performance in two different scenarios and, additionally, regarding generalization and interpretability.
The following experiments are based on both the environment and framework of \cite{Watters2019}, which were extend for our purposes.

\subsection{Environment}
\label{sec:environment}

\begin{figure}
\begin{center}
      \includegraphics[width=0.9\linewidth]{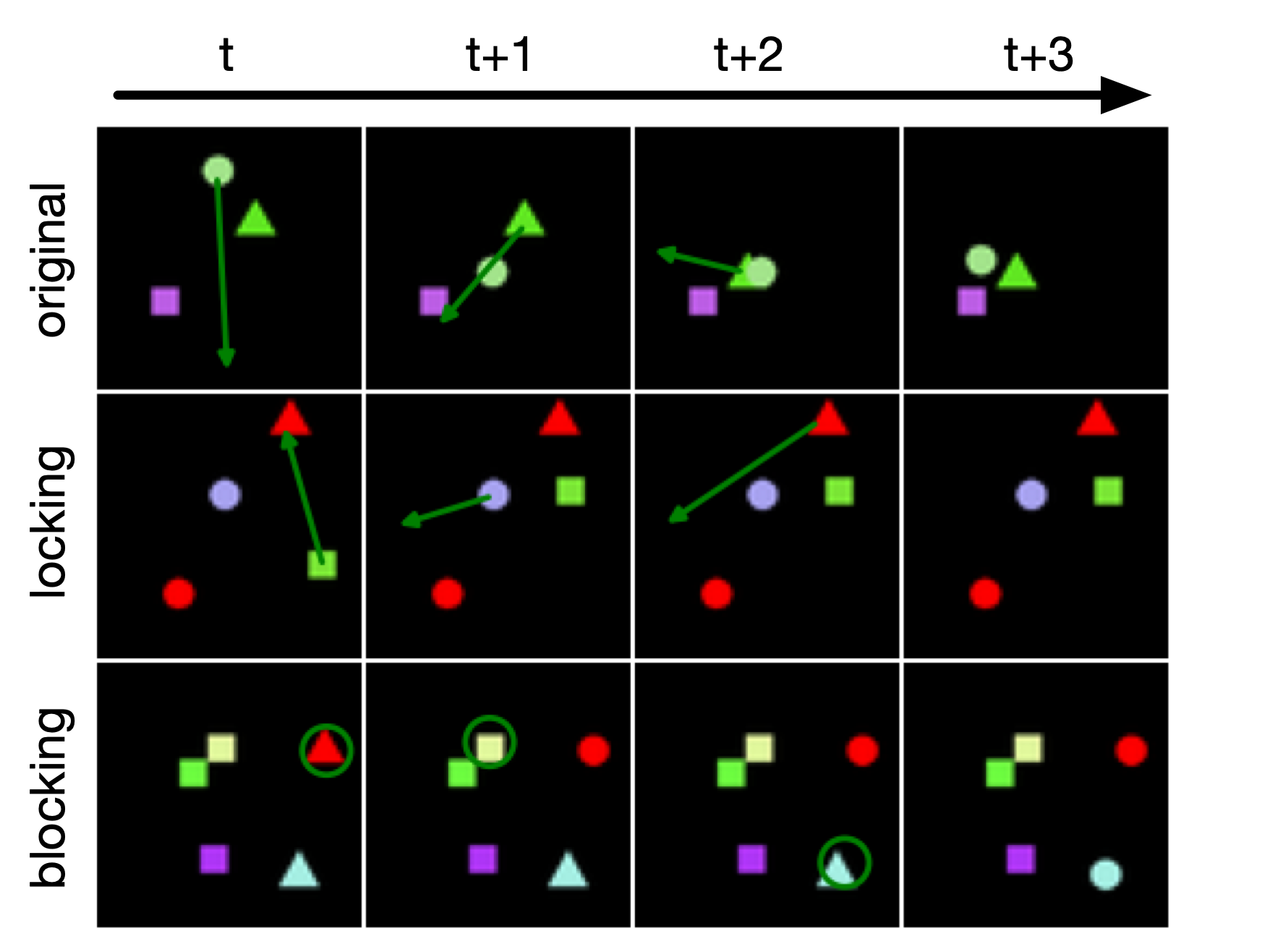}
\end{center}
   \caption{Demonstration of the different mechanics in our environment, given four consecutive actions each (indicated by green arrows (drag action) and circles(click action)). 
   \textbf{Top row:} In the original version \protect\cite{Watters2019}, objects can be moved by dragging them. Dragging moves an object a fixed percentage along the path to the target location (arrow head). We extend this by two new mechanics:
   \textbf{Middle row}: locks (identifiable by red color) cannot be moved and prevent all other objects with the same shape as them from moving.
   \textbf{Bottom row:} when clicking on an object, its shape changes in a predetermined way. However, this is not possible if the clicked object is blocked by touching any other object.}  
\label{fig:env}
\end{figure}

Our environment\footnote[1]{source code available at [github link omitted]} is based on the Spriteworld environment \cite{Watters2019} in which objects can be dragged by clicking on an object followed by a target direction (see Fig. \ref{fig:env}, top row). One can think of this as a robot in a storage facility that first learns how to move objects around before being assigned specific tasks in which that knowledge must be used, e.g.\ sorting specific objects to certain locations.

We extend this environment by adding several new environment mechanics.
First, we add clicking as a new type of object manipulation. By clicking on an object (instead of dragging it), one of its properties (here: shape) will change to the next one from a predetermined list. In the setting of a storage facility robot, this could resemble repacking objects into new containers. 
Second, we define semantic relations between objects. These relations have a discrete effect on object interactions, where the type of effect can be determined by an objects' context. Specifically, we introduce two mechanics based on these semantic relations (see Fig. \ref{fig:env}, middle and bottom row):

\begin{itemize}
\item \textbf{Locking}: If an object is locked, it cannot be moved. For this, we introduce locks to be a specific object type that is characterized by a unique feature (here: red color). Whenever a lock is present anywhere in a scene, it prevents all objects with the same shape from moving. Based on this mechanics, every object can have the semantic states “locked” or “unlocked”, which results implicitly from a discrete property (shape) of other objects in a scene.

\item \textbf{Blocking}: Repacking an object (changing its shape) might require some space. Therefore, if an object touches any other object, its shape cannot be changed. Based on this mechanics, every object can have the semantic states "blocked" or "unblocked", which results implicitly from a continuous property (position) of other objects in a scene.
\end{itemize}

With all mechanics combined, objects will move when dragged (if the are not locked) and change shape when clicked (if they are not blocked). 

\subsection*{Possible Tasks}
Even though the described mechanics are quite simple, they enable the construction of a broad range of diverse tasks. Although we do not explore possible tasks in this paper, we think it is important to emphasize the implications of the chosen mechanics.
Using only the original mechanics (dragging), simple movement tasks are possible, such as goal finding (all objects have to be moved to a specific location) or sorting (objects have to be grouped together based on their color or shape). By including the shape change mechanics (clicking), new solutions are possible. For example, sorting could be easier or more efficient if the agent decides to change the shape of some objects instead of moving them. On the other hand, new tasks can be defined, like gathering a certain number of objects with a specific shape in one place. By additionally including the locking and blocking mechanics, complex puzzle-like environments can be created. For example, a scenario where the agent has to move blocking objects first in order to access a lock that prevents a target object from being manipulated.
In that sense, the described environment captures some of the core mechanics in ATARI games, where relational concepts like key-lock relations or spatial closeness are common elements. Even though the state spaces for solving tasks in such environments might be very diverse, they could be constructed from the same set of task-independent elements, including the perceptual and semantic state of an object. Learning to represent such elements in a factorized manner will make such tasks accessible, while allowing the task training to focus on task-specific knowledge (e.g.\ goals, conditions and order of object manipulations).

\hfill \linebreak
In our implementation, objects are represented as a feature vector $z^k \in [0,1]^{8}$ with two continuous features for position and three features for both shape and color respectively. We uniformly sample 27 discrete possible shapes with $z_{shape}^k \in \{0,0.5,1\}^{3}$. The three continuous color dimensions correspond to the HSV color space. While lock objects always have a pure red color, regular objects may take any other hue that is not too close to red. As in the original environment by \cite{Watters2019}, actions are represented as $a \in [0,1]^{4}$ containing the coordinates two consecutive clicks.

In the following sections, we consider two specific environments created with the described mechanics.

\subsection{Training}
\label{sec:training}

Using \ac{rl}, \cite{Watters2019} have shown that an object-specific \ac{tm} can be trained with intrinsic motivation (curiosity) and later be used to learn a range of tasks. Since we are only focused on a very specific part, we can simplify certain aspects of their setup to investigate our model more effectively without loss of generality: 

\begin{itemize}
\item \textbf{Object Extractor:} 
An object extractor like MONET \cite{Burgess} is capable of representing objects with a small set of disentangled features and can be trained independently in a self-supervised manner \cite{Watters2019}. Since the specifics of the object representations are not relevant in our case, we skip the object extractor and provide an object list directly. This also simplifies the training compared to \cite{Watters2019} because MONET might change the order of objects across time steps.

\item \textbf{Data Acquisition:}
\acp{tm} are usually trained self-supervised on observed environment transitions.
As \cite{Pathak2017,Watters2019} have shown, an exploration policy with intrinsic motivation like curiosity can be used to gather meaningful transitions. We simplify this by replacing the exploration policy with a very basic action sampling function. However, meaningful transitions are still very sparse in our data set, as described in Section \ref{sec:minimal} and \ref{sec:multirelation} 

\item \textbf{Tasks:}
With a sufficiently accurate \ac{tm}, tasks can be learned purely on the \ac{tm}. In such a case, the performance on possible tasks is limited by both the type of mechanics learned by the \ac{tm} and the quality of its predictions. Therefore, we focus purely on the performance of the \ac{tm} as a proxy for the upper limit of possible task performance.
\end{itemize}

Considering these simplifications, we train all models supervised on batches of environment transitions $[( z_t, a_t),  z_{t+1}]$ with a batch size of 10 using plain Adam (default parameters) and L2 loss. Importantly, these transitions are not a static dataset but rather dynamically collected from the environment in trajectories of length 10 utilizing a simple random action sampling function. The specific action sampling functions are discussed in more detail in sections \ref{sec:minimal} and \ref{sec:multirelation}.

\subsection{Measures}

In our environment, objects may change certain properties ($z_{p}^k, p \in \{shape, position\}$) in response to actions. An object's property is considered to be predicted correctly, if its prediction differs less than a threshold $th_\Delta = 0.05$  from the ground truth. With this, the percentage of correct predictions for specific properties $PC_{p}$ can be used as a measure of performance. In order to assess a model more precisely, we consider two aspects for evaluation. For semantics, we evaluate \textbf{whether} a change is predicted at all: 

\begin{equation}
\Delta z_{p}^k > th_\Delta.
\end{equation}

For dynamics, we evaluate if the the \textbf{correct} change is predicted:

\begin{equation}
\Delta z_{p}^k - \Delta z_{p, GT}^k < th_\Delta.
\end{equation} Still, this does not give enough insight into the learned knowledge. Therefore, we evaluate these two measures for different object states individually. For locked objects, we measure the percentage of objects that were not moved. For unlocked objects, we measure the percentage of objects that were moved the correct distance (similarly for blocked and unblocked objects). 

Note that the objects semantic states are not explicitly contained in our training data and have to be inferred by the model. However, since we have access to the data generation process, we can use this knowledge for evaluation.

Finally, we determine the probabilities of each object state empirically by simply counting their occurrences across $10^6$ randomly generated scenes as a measure of sparsity.

\subsection{Minimal Environment}
\label{sec:minimal}

We first test the design of our model with the most simple environment extension, including only the locking mechanics. As illustrated in Fig. \ref{fig:env}, middle row, objects can be moved by dragging if no lock with the same shape is present. In this environment, a purely object-specific \ac{tm} will not be able to reliably predict an object's future state because other objects (such as locks) have to be taken int account.

A scene $\bm{z} \in \mathbb{R}^{5\times 8}$ consists of 0-2 lock objects and 1-3 regular objects, all of which can have one of three possible shapes. The actions are generated by randomly choosing one object and sampling a drag action, constructed from a random click position within that object followed by a random movement direction. In each time step, exactly one object is interacted with and the probability for a interacted object to be locked is approximately 42\%. Since the \acl{tm} $T_{Obj}$ predicts all 5 object slots individually at each time step, the probability for it to encounter a locked object is 8.5\%

We configure $S$ to predict a single additional value for each object which can be used to encode the locking state. For the internal (A) and sequential (B) models, the output of $S$ is therefore $\hat{\bm{z}} \in \mathbb{R}^{5\times 9}$. Independent of the perceptual object representation's compactness, this ensures that all of the $M=8$ perceptual features can theoretically be preserved while additionally containing the locked state. For the parallel model (C), the output of $S$ is $\hat{\bm{z}} \in \mathbb{R}^{5\times 1}$ and concatenated to the perceptual object representation $\bm{z} \in \mathbb{R}^{5\times 8}$.

\subsubsection*{Results}

\begin{figure}
\begin{center}
      \includegraphics[width=.9\linewidth]{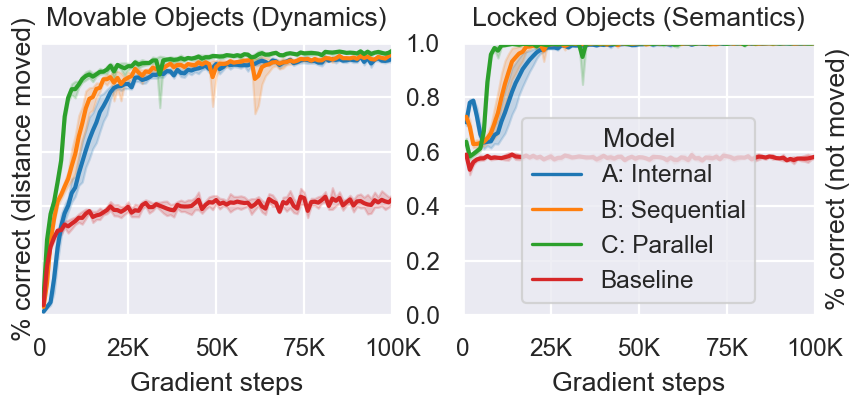}
\end{center}
   \caption{Training performance on the minimal task (Section \ref{sec:minimal}, averaged across 5 runs). While a purely object-wise baseline (without semantic module) fails to solve the task, all of the proposed models perform similarly well. 
      \textbf{Left} In case an object changes position, correctly predicting the amount of change for that object.
   \textbf{Right}: If an object is locked, correctly predicting that it will not change its position. This also includes locks themselves ($\sim 50\%$ of locked objects). Note that the parallel model learns roughly twice as fast as the parallel model.
    }
\label{fig:minimal_env}
\end{figure}

The results shown in Fig. \ref{fig:minimal_env} confirm that our proposed semantic module $S$ is able to consider inter-object relations in a way that enables a single object-specific \ac{tm} to correctly predict future object states, which fails for a baseline model without $S$. Furthermore, a more explicit separation does not decrease the performance, and instead even increases the training speed noticeably. This supports our hypothesis that the inductive bias can speed up training by imposing structure that otherwise needs to be learned implicitly.

\subsection{Multi-Relation Environment}
\label{sec:multirelation}

We now consider a more complex variant of our environment with multiple independent relations. For this, we extend the previous experiment with the click action to change shapes (see Fig. \ref{fig:env}, bottom row). Now, objects can be locked based on their shape and blocked based on their spatial proximity to other objects. Additionally, we increase the number of possible shapes which reduces the chance of the locked case (i.e. both an object and a lock with the same shape occurring in one scene).

A scene $\bm{z} \in \mathbb{R}^{7\times 8}$ consists of 0-2 lock objects and 1-5 regular objects, all of which can have one of five possible shapes. The actions are generated by randomly choosing one object and sampling either a drag or click action with equal probability. An action is constructed from a random click position within that object followed by a random movement direction for dragging or a second random point within the object for clicking. In each time step, exactly one object is interacted with and the probability for a selected object to be locked or blocked is approximately 38\% and 22\% respectively. Considering that the transition model $T_{Obj}$ predicts all 7 object slots independently in each time step, the probability for it to encounter a locked or blocked object is 5\% and 3\% respectively. However, since the effect can only be observed under the corresponding action, the probabilities reduce even further to 2.5\% and 1.5\%. For this reason, this scenario is especially difficult because the relevant transitions are very sparse.

In the minimal experiment (Section \ref{sec:minimal}), we configured $S$ to predict a single additional value, exploiting our knowledge about the environment. In order to show that the proposed models also work in the absence of such designer knowledge, we configure them to predict eight additional values, even though two would be sufficient. For the internal (A) and sequential (B) model variants, the output of $S$ is therefore $\hat{\bm{z}} \in \mathbb{R}^{7\times 16}$. For the parallel model variant (C), the output of S is $\hat{\bm{z}} \in \mathbb{R}^{7\times 8}$ and concatenated to the perceptual object representation $\bm{z} \in \mathbb{R}^{7\times 8}$.

\subsubsection*{Results}

\begin{figure}
\begin{center}
      \includegraphics[width=0.95\linewidth]{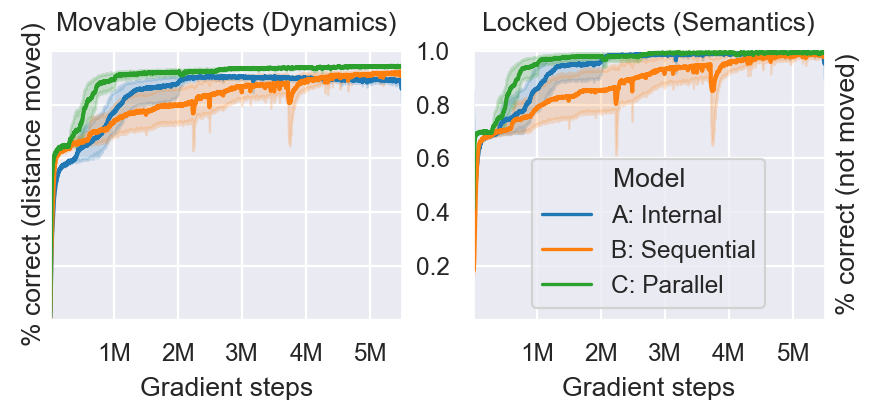}
\end{center}
   \caption{Training performance on the multi-relation task (Section \ref{sec:multirelation}, averaged across 10 runs). Due to increased difficulty, differences between models become more pronounced. While the parallel variant learns almost twice as fast as the internal, the sequential variant struggles much more with the sparsity of meaningful transitions. Its shallow average learning curve is due to high variance across individual runs.
}
\label{fig:multi_env}
\end{figure}

With the increased difficulty, the models now differ noticeably in their learning speed regarding movable and locked objects (see Fig. \ref{fig:multi_env}). Given the sparsity of meaningful transitions, the training time generally increased by an order of magnitude compared to the simpler environment. Nevertheless, the parallel model still shows the improvement in training efficiency, again supporting our hypothesis from Section \ref{sec:model} (Model learning). The sequential model, on the other hand, shows a higher variance between runs and is therefore slower on average. We suspect this due to the fact that the perceptual features are not guaranteed to be preserved in this model variant, possibly resulting in less training stability.

\begin{figure}
\begin{center}
		\includegraphics[width=0.95\linewidth]{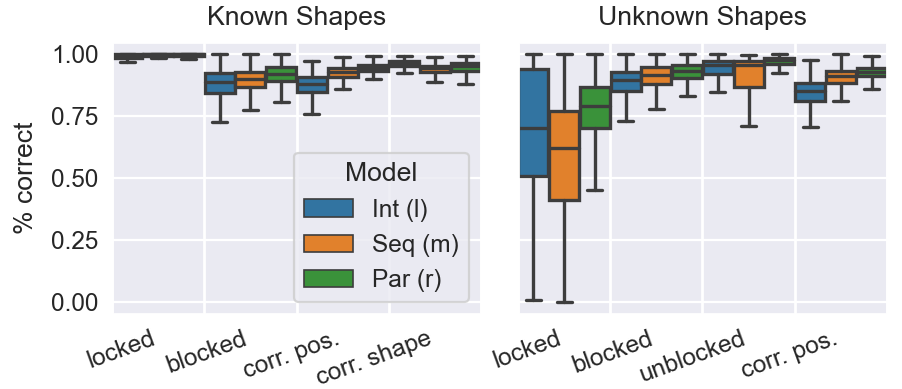}
\end{center}
   \caption{Performance after training on the multi-relation task (Section \ref{sec:multirelation}, measured across 10 runs over $10^6$ scenes) for known training objects \textbf{(left)} and for generalization to previously unseen objects \textbf{(right)}. The measures locked and blocked concern the detection of semantic states, while the remaining measures concern dynamics prediction. All models perform comparably well on training data, but differ significantly for generalization to new objects. Generally, the parallel model (Par) with the most explicit knowledge separation performs best.
   }
\label{fig:generalzation}
\end{figure}

The final performance after $5.5 \times 10^6$ gradient steps, evaluated across 1,000 batches with 1,000 scenes each, is shown in Fig. \ref{fig:generalzation} (left). Regarding semantic knowledge, the detection of blocked objects seems to be slightly more difficult in general than detecting locked objects with $PC_{blocked} \approx 0.8 < PC_{locked} = 1$ for all models. We suppose this has two reasons: first, the transitions are significantly more sparse for blocking compared to locking. Second, the blocked state depends on spatial proximity. Since this is a continuous relation for which the exact threshold needs to be learned, it is a more difficult problem than matching a discrete shape. 
Nevertheless, even though the blocked state cannot be predicted as reliably, the dynamics prediction for unblocked objects is better than for unlocked objects, regardless of the architecture ($PC_{corr. shape} > PC_{corr. pos.}$). Again, we suspect this is because the correct shape is a discrete property while the correct position is continuous.

In general, a slight trend can be noticed across all measures in which the performance increases from internal to sequential to parallel, with the latter always performing best. This is most notably for predicting the dynamics of unlocked objects ($PC_{corr. pos.}$). Even though the effect is less pronounced, it suggests a correlation between performance and the degree of explicitness in knowledge separation. In total, the parallel model with the most explicit separation performs best with lower variances.

The results also show that the semantic module is robust regarding the dimensionality of its produced embedding. Given that it is sufficiently large, increasing the number of output features has no significant effect on performance and training time. Preliminary experiments further solidify this observation.

\subsection{Generalization}
\label{sec:generalization}

We now test whether the proposed inductive bias improves generalization. If semantic relations are disentangled from perceptual features and environment dynamics, they should be transferable to new objects. For example, a generic concept of "blocking" or "locking" is independent of specific object shapes and should therefore generalize to new shapes.

For this, we use the previously trained models as described in Section \ref{sec:multirelation} and evaluate their performance in a new environment with previously unseen objects. During training, objects could have one of five shapes. Here, objects have one of 22 unseen shapes.
As in Section \ref{sec:multirelation}, we evaluated all models for 1,000 batches, each batch containing 1,000 scenes constructed with 5 randomly chosen evaluation shapes per batch.

Importantly, the next shape for a clicked object cannot be determined from the input, but a fixed order of shapes has to be memorized instead. The \ac{tm} is therefore now unable to predict the correct next shape for unseen objects. In effect, the dynamics now change even though semantics do not. However, if dynamics and semantics are separated, it would be still possible to predict whether an object is locked or not. Additionally, learning a new (or adapting the old) dynamics model leveraging the already learned semantics knowledge could be much easier than learning from scratch. 
Since the "correct shape" measure is now unsuited to evaluate the performance in a meaningful way, we define the "unblocked" measure that counts any change of shape as correct for unblocked objects.

\subsubsection*{Results}
The results in Fig. \ref{fig:generalzation} (right) show drastic differences between the model variants.
For both the internal and sequential model, the prediction of locked objects is much less reliable, often times failing completely. We further noticed that their performance varies greatly across different unseen regions in the shape space, and the location of those regions varies across multiple runs. This indicates a lack of disentanglement between perceptual features (here: shape) and semantic knowledge for both model variants, resulting in poor generalization. 

The performance of the parallel model variant, on the other hand, is more robust and generalizes better to unseen objects. With only five training shapes, it has learned a rather generic concept of locking that is less tied to specific shapes. Of curse, with a sufficiently large number of training shapes, all models will eventually be able to equally generalize to unseen shapes. However, providing the complete range of possible feature characteristics (and their combinations) during training is often times not feasible - even if the training is simulated. Returning to our example of a storage facility robot from Section \ref{sec:environment}, a robot trying to learn generic object-related concepts may not have access to all possible objects during training. Instead, it should be able to learn by interacting with a very limited set of practice objects. In scenarios like this, overfitting to the training distribution is a typical problem. Our approach addresses this by encouraging knowledge separation and therefore disentanglement between perception, semantics and dynamics. This separation also enables the possibility to reuse semantic knowledge to train specific task policies later on. We see this disentanglement as one of the main benefits of our approach and examine this in more detail in Section \ref{sec:interpretability}. 

In summary, our results suggest that the degree of explicitness in knowledge separation does indeed correlate with the ability to effectively generalize to new objects. Therefore, our proposed inductive bias could be a promising approach towards generic object centered concepts.

\subsection{Disentanglement and Interpretability}
\label{sec:interpretability}

\begin{figure}
\begin{center}
      \includegraphics[width=.95\linewidth]{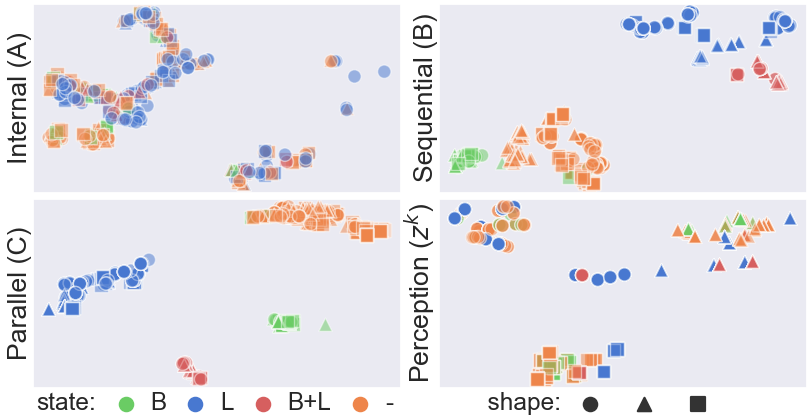}
\end{center}
   \caption{t-SNE visualization of the embeddings learned by the semantic module in all three configurations, for all possible combination of semantic states (\textbf{B}: blocked, \textbf{L}: locked, \textbf{B+L}: both, \textbf{-}:none). While a separation of semantic concepts is not possible in model A, models B and C produce distinct clusters. However, the clustering of C provides the most generic separation that is indifferent to specific shapes.
   }
\label{fig:embedding}
\end{figure}

In order to further assess the knowledge separation, we visualize the embeddings $\hat{z}^k$ produced by the semantic module $S$ in all three configurations in Fig. \ref{fig:embedding}. It clearly shows a varying degree of disentanglement between semantic and perceptual information: While the internal model does not form semantic clusters at all, the other two models clearly do. However, the clusters formed by the sequential model are not as cleanly separated and seem to be entangled with perceptual information, noticeable by the grouping of shapes inside the semantic clusters. Both aspects are undesirable for robust generic concepts that aim to be indifferent to specific object types, especially since the perceptual information (shape, position, color) can be already extracted from the perceptual object representations (see Fig. \ref{fig:embedding}, $z^k$). Consequently, the parallel model provides the most clear, disentangled and generalizing semantic representation. Furthermore, the clusters of C  could be easily mapped to shape independent semantic labels (locked, blocked). This would greatly improve transparency by providing a method for humans to easily understand the agents semantic perception of the world and, consequently, its basis for action.

\section{Conclusion and Future Work}

We proposed a novel inductive bias for model-based RL that explicitly separates object-based knowledge into semantics and dynamics. It follows the assumption that objects can have semantic states which are not deducible from its perceptual features alone. Instead, such states are dependent on an objects' context and are noticeable by their influence on interaction dynamics.
We implemented this bias with a dedicated semantic module. This module can be integrated with an object-wise \ac{tm} in three different ways, which vary in their degree of explicitness of semantic separation. Within a newly introduced environment extension, our experiments show that increasing the explicitness of this separation correlates with multiple desirable qualities. 

First and most notably, it facilitates generalization to new objects by disentangling semantic knowledge from perceptual features and dynamics knowledge. Consequently, only a small number of training objects is necessary to learn generic semantic knowledge that can be applied to unseen objects. Additionally, this knowledge is then less susceptible to changes in dynamics.
Second, this disentanglement also improves interpretability.  
Third, it can mitigate sparseness of meaningful transitions in training data, resulting in increasing sample efficiency and decreasing training time. In our experiments, the degree of explicitness in separation also correlates with a slight increase in performance.
Across all measures, our parallel model variant (which implements the most explicit knowledge separation) performs best.

With these promising results as groundwork, the presented approach can be a mechanism for learning emergent generic concepts that can be reused across multiple tasks. The next step into this direction is to investigate the  transfer of semantic knowledge to novel tasks. Focusing more on the emergence, another valuable direction is to embed our approach into an end-to-end RL framework with curiosity driven exploration \cite{Pathak2017,Watters2019}, which could allow the model to freely learn the semantic concepts it values the most.

While these findings are currently limited to object-wise \acp{tm}, transferring the inductive bias to more complex \acp{tm} (e.g\ including physics \cite{Janner2019}) provides an interesting challenge for future work.
Finally, different implementations for a semantic module such as PrediNet \cite{Shanahan2020} can be explored. While this might allow more powerful semantic concepts, it also poses an opportunity to substantiate a symbolic link towards the learned semantic relations, aiming to bridge the gap between bottom up learned representations and human concepts.

\bibliographystyle{IEEEtran}
\bibliography{RelToPred}

\end{document}